\documentclass[sigconf,natbib=true]{acmart} %

\usepackage{lipsum}

\AtBeginDocument{%
  \providecommand\BibTeX{{%
    \normalfont B\kern-0.5em{\scshape i\kern-0.25em b}\kern-0.8em\TeX}}}

\copyrightyear{2023}
\acmYear{2023}
\acmConference[Gen-IR@SIGIR2023]{The First Workshop on Generative Information Retrieval}{July 27, 2023}{Taipei, Taiwan}
\acmBooktitle{The First Workshop on Generative Information Retrieval (Gen-IR@SIGIR23), July 27, 2023, Taipei, Taiwan}
\acmDOI{}
\acmPrice{}
\acmISBN{}
\setcopyright{rightsretained}

\usepackage{xspace}
\usepackage{algpseudocode}
\usepackage{enumitem}
\usepackage{soul}
\usepackage{color,soul}

\newcommand\rouge{\textsc{Rouge}\xspace}
\newcommand\bart{\textsc{Bart}\xspace}
\newcommand\segenc{\textsc{SegEnc}\xspace}
\newcommand\ro{\textsc{Rouge-1}\xspace}
\newcommand\rt{\textsc{Rouge-2}\xspace}
\newcommand\rl{\textsc{Rouge-L}\xspace}
\newcommand\socratic{\textsc{Socratic}\xspace}
\newcommand\pegasus{\textsc{Pegasus}\xspace}
\newcommand\ours{\textsc{Qontsum}\xspace}

\begin{document}

\title{\ours: On Contrasting Salient Content for Query-focused Summarization}

\author{Sajad Sotudeh}
\email{sajad@ir.cs.georgetown.edu}
\affiliation{%
  \institution{IR Lab, Georgetown Univeristy}
  \country{Washington D.C., USA}
}

\author{Nazli Goharian}
\email{nazli@ir.cs.georgetown.edu}
\affiliation{%
  \institution{IR Lab, Georgetown Univeristy}
  \country{Washington D.C., USA}
}


\begin{abstract}
Query-focused summarization (QFS) is a challenging task in natural language processing that generates summaries to address specific queries. The broader field of Generative Information Retrieval (Gen-IR) aims to revolutionize information extraction from vast document corpora through generative approaches, encompassing Generative Document Retrieval (GDR) and Grounded Answer Retrieval (GAR). This paper highlights the role of QFS in Grounded Answer Generation (GAR), a key subdomain of Gen-IR that produces human-readable answers in direct correspondence with queries, grounded in relevant documents. In this study, we propose \ours, a novel approach for QFS that leverages contrastive learning to help the model attend to the most relevant regions of the input document. We evaluate our approach on a couple of benchmark datasets for QFS and demonstrate that it either outperforms existing state-of-the-art or exhibits a comparable performance with considerably reduced computational cost through enhancements in the fine-tuning stage, rather than relying on large-scale pre-training experiments, which is the focus of current SOTA. Moreover, we conducted a human study and identified  improvements in the relevance of generated summaries to the posed queries without compromising fluency. We further conduct an error analysis study to understand our model's limitations and propose avenues for future research.
\end{abstract}

\maketitle

\section{Introduction}

In recent years, the proliferation of digital content has led to an explosion of data, making it increasingly challenging to extract relevant information and insights from large amounts of text. Summarization, the process of condensing large volumes of text into shorter, more manageable summaries, has emerged as a promising solution to this problem. Among different types of summarization, query-focused summarization (QFS)~\cite{Dang2005OverviewOD, Daum2006BayesianQS,Katragadda2009QueryFocusedSO,Badrinath2011ImprovingQF,xu-lapata-2020-coarse,Xu2022DocumentSW,Laskar2022QFS,Yuan2022FewshotQS} has gained significant attention due to its ability to generate summaries tailored to specific user queries or information needs.

{
QFS is situated within the larger domain of Generative Information Retrieval (Gen-IR), an area that seeks to revolutionize information extraction from vast document corpora by employing generative techniques. In order to understand the relationship between QFS and Gen-IR, it is essential to first explore the divisions within Gen-IR. Generative Information Retrieval can be divided into two main subfields: Generative Document Retrieval (GDR) and Grounded Answer Generation (GAR). GDR is concerned with retrieving a ranked list of documents w.r.t a given query, while GAR focuses on generating specific answers, grounded on relevant documents~\footnote{{Definitions of Gen-IR, GDR, and GAR are taken from the Gen-IR workshop at} \url{https://coda.io/@sigir/gen-ir}}. In this context, QFS can be viewed as a method associated with GAR.} QFS is essential in a wide range of natural language processing applications, such as information retrieval~\cite{Metzler2007SimilarityMF, Yih2007ImprovingSM, Gu2016AFA,Chen2020AbstractiveSG,Althammer2021DoSSIERCOLIEE2L,Choongwon2022QFS}, and document analysis~\cite{GoldsteinStewart1999SummarizingTD,Lin2003AutomaticEO,Gupta2007MeasuringIA}. In information retrieval, for example, QFS can help users quickly and effectively identify relevant information from large volumes of search results. Similarly, in document analysis, query-focused summarization can provide decision-makers with key insights and information necessary for making informed decisions~\cite{Hsu2021DecisionFocusedS}.

{
By condensing large volumes of information into a focused summary that directly addresses the user's information need, QFS enhances the informativeness of generated responses. This ability is particularly beneficial for tasks such as question answering, chatbots, and personal assistants, where the quality and relevance of generated responses are critical for user satisfaction and engagement. Incorporating QFS into Gen-IR systems can help address the issue of generating excessively long or irrelevant responses, a common challenge in natural language response generation. By providing users with more concise and targeted information, QFS can improve the  effectiveness of Gen-IR systems, ultimately enhancing the user experience.}

\begin{figure}
    \centering
    \includegraphics[scale=0.55]{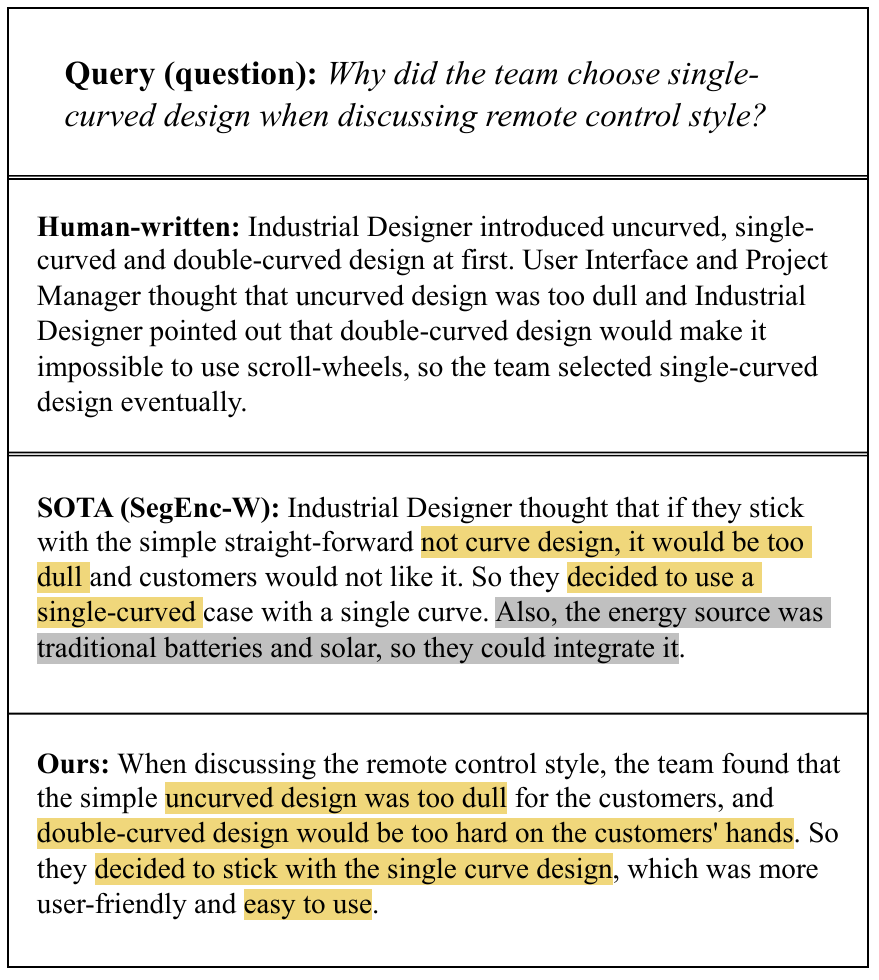}
    \caption{A sample query/question with its human-written and system-generated responses (i.e., summaries) from the QMSum benchmark. Yellow highlights in the summaries signify text spans that are directly relevant to the query. In contrast, gray highlights denote spans that are not relevant to the query, as produced by the state-of-the-art (SOTA) summarizer. Relevance is evaluated based on the system's ability to generate \textit{query-related} responses, without being limited to solely the information in the human-written summary.
    }
    \label{fig:intro_fig}
    \vspace{-2em}
\end{figure}

Existing methods for query-focused summarization can be broadly categorized into extractive, abstractive, and hybrid approaches. Extractive approaches involve selecting and aggregating the most important information from the input document based on various heuristics or statistical models. Abstractive approaches, on the other hand, aim to generate a summary that captures the essence of the input document in a more human-like manner, often by generating novel phrases or sentences. Hybrid approaches aim to combine the strengths of both extractive and abstractive methods. Despite significant progress in recent years, these approaches still face several challenges, such as the difficulty of capturing the nuances of human language and understanding the contextual information in the input document. Moreover, ensuring the faithfulness and relevance of generated summaries remains a critical challenge, particularly in scenarios where the input document contains complex information or where the query is granular. In this paper, we focus on enhancing the relevance of the generated summary to the query, a challenge that the state-of-the-art systems often encounter, {as shown in Figure }\ref{fig:intro_fig}.

In the context of an ever-growing volume of information, long-input summarization has become increasingly important, as it can facilitate the efficient and effective extraction of key insights from extensive documents. Thus, our study opts to focus on long-input QFS, rather than short-input QFS. To address the challenges and limitations posed by current techniques in the domain of query-focused summarization, this study introduces an innovative method that employs contrastive learning~\cite{Chopra2005LearningAS} to distinguish between pertinent and non-pertinent content within the input document. The proposed methodology seeks to incorporate the most salient spans of the source document into the summarization process by juxtaposing them against negative regions that are typically the focus of attention for the summarization system. We posit that the incorporation of contrastive learning into query-focused summarization can bolster the model's capacity to discern relevant information from input documents, thereby yielding more relevant and pertinent summaries. Our empirical findings, based on two long-input datasets, demonstrate either enhanced or comparable performance in relation to the previous state-of-the-art, while simultaneously reducing computational cost.  
In summary, this investigation presents a promising new avenue for augmenting the relevance of query-focused summarization and enriching the quality of generated responses in Generative Information Retrieval and Query-focused Summarization. Our contributions are twofold:


\begin{itemize}
    \item We present a novel QFS system that combines contrastive learning with state-of-the-art techniques, surpassing or achieving the performance of existing approaches on relevant benchmark datasets.
    
    
    \item We conduct an evaluation of our model, including automatic performance comparisons, human assessment, and error analyses, to gain insights into the model's strengths, limitations, and potential future research directions.

\end{itemize}



\section{Related work}

Query-focused summarization (QFS) is a specialized area within automatic text summarization, where the objective is to generate summaries tailored to address a specific query by selecting and condensing relevant information from a given document collection. In the early stages of QFS research, extractive methods were the predominant techniques used to generate summaries. These methods identify salient sentences or passages from the input documents and combine them to form a summary without altering the original text. One popular approach is query expansion, which enriches the user query with additional terms to better capture the user's intent and improve the relevance of the extracted summaries~\cite{Metzler2007SimilarityMF,Sriram2010ShortTC}. Another method is query-biased summarization in information retrieval, which ranks sentences based on their similarity to the user query and their importance within the document~\cite{Tombros1998AdvantagesOQ,Metzler2008MachineLS,Gu2012TowardsES}. For instance, Graph-based methods, such as LexRank~\cite{Erkan2004LexRankGL} and TextRank~\cite{Mihalcea2004TextRankBO}, represent documents as graphs, where nodes correspond to sentences and edges indicate the similarity between them. By analyzing the graph structure, these methods identify and extract key sentences based on their centrality and relevance to the user query.

The field of query-focused summarization (QFS) has experienced significant advancements due to the incorporation of deep learning and neural networks. Abstractive methods have become increasingly popular in QFS research, as they generate flexible and coherent summaries by paraphrasing and rephrasing the input text. Various techniques, such as sequence-to-sequence models~\cite{Sutskever2014SequenceTS}, attention mechanisms~\cite{Cho2014LearningPR}, and transformer-based architectures~\cite{Vaswani2017Att}, have been employed to improve QFS model performance by assigning weights to input tokens. Pre-trained language models, like Bart~\cite{Lewis2020BARTDS}, and transformer-based architectures form the foundation of state-of-the-art (SOTA) models for QFS. 

The availability of high-quality query-focused summarization datasets, such as QMSum~\cite{zhong2021qmsum} and SQuALITY~\cite{wang-etal-2022-squality}, has fueled increased interest in QFS. Recent research has explored extract-then-generate methods, including passage/answer retrieval~\cite{Baumel2018QueryFA,Laskar2020QueryFA,Su2021ImproveQF,zhong2021qmsum,wang-etal-2022-squality}. Other methods involve adapting attention mechanisms to query-focused summarization through query-utterance interactions~\cite{Liu2023TUQFS}, adapting SOTA summarizers like \bart to long-input QFS through overlapping segment-based summarization~\cite{Vig2022SEGENC}, and replacing full attention with block-wise attention~\cite{Beltagy2020LongformerTL}. Additional approaches include generating pseudo queries for ranking evidence sentences~\cite{Xu2020GeneratingQF}, employing data augmentation techniques~\cite{pasunuru2021data}, and pre-training language models with dialog-specific~\cite{Zhong2021DialogLMPM} and question-driven~\cite{Pagnoni2022SocraticPQ} objectives. These advancements in QFS research have contributed to the development of more effective and coherent summarization models, paving the way for further innovations in natural language understanding and generation.

Different than prior work, our approach incorporates contrastive learning to help the model perform generation from relevant regions of the input document, with the goal of increasing the summary's relevance to the query. To the best of our knowledge, this is the first attempt to utilize contrastive learning in QFS. By leveraging this technique, we aim to guide the model to focus on the most relevant parts of the input document, thereby producing summaries that are better aligned with the query.

    



\begin{figure*}
    \centering
    \includegraphics[scale=0.46]{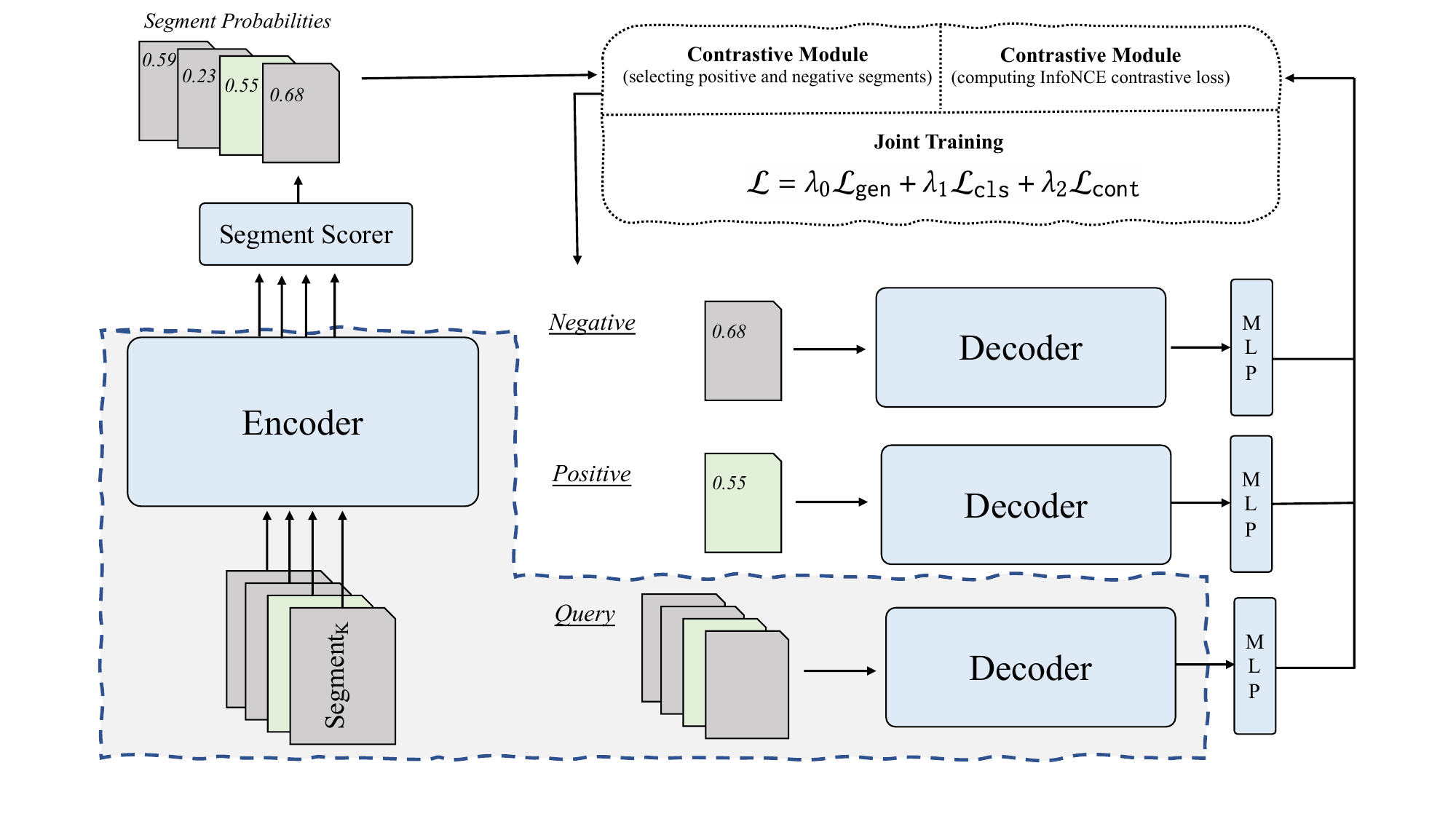}
    \caption{Overview of \ours model. The schema inside the dashed line represents the \segenc baseline, which performs segment-based encoding, concatenates the encoding into a dense embedding space, and utilizes a decoder to attend to this embedding. Our model builds upon the \segenc baseline: we first segment the input documents into fixed-length overlapping segments and then feed these segments into the Encoder network. The Segment Scorer, placed on top of the encoder, pools the segment representations and projects them into segment probabilities. Given the probabilities of the segments and the gold segments (in green), we generate positive and negative samples for contrastive learning. Both types of samples are fed into a shared-weight decoder along with the query, enabling the summarization model to learn to focus on relevant segments. The entire network is finally optimized using three losses: generation, classification, and contrastive which are all combined with a balancing $\lambda$ parameter.}
    \label{fig:fig_model}
\end{figure*}

\section{Background: Segment-based long summarization}

Prior to presenting our contrastive learning approach, we present an overview of the SOTA Segment Encoder (\segenc) summarization model, as proposed by \citet{Vig2022SEGENC}, which serves as the backbone summarization component of our framework. In the \segenc model, the source document is initially divided into overlapping segments of fixed length~\footnote{We use 512 tokens, with each segment exhibiting a 50\% overlap with its adjacent segment (empirically determined).}, each of which is appended to the query and subsequently encoded using a conventional Transformer model, such as \bart. In this context, the encoder focuses on both query and segment tokens through its self-attention mechanism \cite{Vaswani2017Att}, and generates token representations for each segment that are attuned to the query. After processing all segments with the shared encoder, their representations are merged into a single continuous embedding sequence, which is then fed into the decoder to create a summary tailored to the query (i.e., the response). Owing to the absence of cross-attention between encoded segments, the attention mechanism's scalability is linearly proportional to the number of segments and, consequently, the length of the input document. Nevertheless, the decoder is capable of attending to all encoded segments jointly, allowing the summarizer to operate in an end-to-end manner. 

The generation loss for this network is calculated using cross-entropy, which measures the difference between the predicted summary output and gold summary as follows: 

\begin{equation}
    \mathcal{L}_{\texttt{gen}} = -\sum_{t=1}^{T} \log p(y_t|y_{<t},x)
\end{equation}
where $T$ is the length of the generated summary, $y_t$ is the $t$-th token in the generated summary, $y_{<t}$ represents the previously generated tokens, and $x$ is the input text. The generation loss measures the negative log-likelihood of generating the ground truth summary given the input text, and the goal during training is to minimize this loss.

\section{Methodology}

This section introduces our proposed approach for query-focused summarization using contrastive learning, which involves applying Generative Information Retrieval techniques to the task of text summarization. The key aspect of our approach is based on the concept of contrastive learning, which has been successfully applied in various other domains~\cite{Hoffer2014DeepML,Bell2015LearningVS,Khosla2020SupervisedCL, Gao2021SimCSESC}. Our approach, named \underline{\textbf{Q}}uery-focused C\underline{\textbf{ont}}rastive \underline{\textbf{Sum}}mmarization (i.e., \ours), leverages contrastive learning to distinguish between positive and negative instances during training, enabling discrimination between important (positive) and unimportant (negative) content in the summarization process. Figure \ref{fig:fig_model} provides a detailed illustration of our model. I the following sections, we provide a systematic breakdown of our proposed approach. 


\subsection{Segment Scoring}

Our architecture incorporates a feed-forward neural network as a critical component for scoring encoded segments in an extractive framework. This network plays a fundamental role in dynamically selecting negative samples during the training process. Specifically, the encoded segment representations, which are associated with the \textsc{<s>} representations as the classification head (i.e., [\texttt{CLS}]), are fed into a feed-forward neural network with a Sigmoid classifier.

\begin{equation}
    p_i = \sigma (h_i W_i + b_i)
\end{equation}
in which $p_i$ represents the extraction probability of the $i$-th segment, $\sigma$ is the Sigmoid activation function, and $W_i$ and $b_i$ are the learnable parameters. After obtaining the segment probabilities, we minimize the cross-entropy function as the classification loss, given the ground-truth labels for each segment:

\begin{equation}
\mathcal{L}_{\texttt{cls}} = - \sum_{i=1}^{S} y_{i} \log(p_{i}) + (1 - y_{i}) \log(1 - p_{i})
\end{equation}
where $S$ is the number of segments, $y_i$ is a binary label indicating whether the $i$-th segment is contributive in summary or not, and $\mathcal{L}_{\texttt{cls}}$ is the classification loss. 

\subsection{Contrastive Learning Framework}
In this subsection, we describe the contrastive learning framework that forms the basis of our proposed approach to query-focused summarization. The main objective of contrastive learning is to distinguish between similar (positive) and dissimilar (negative) instances by learning to map them to different regions in the embedding space. Our framework selects positive (fixed) and negative (dynamically) instances during the training process, which allows the model to learn effective representations that can discriminate between important and unimportant content for summarization.

\subsubsection{Positive and Negative instance selection}
To generate positive instances, we first check if ground-truth labels are available in the dataset. If not, we generate supervised labels, which will be further explained in Section \ref{sec:setup}. We examine each segment $S_i$ to determine if it contains any gold spans or sentences. If it does, we add the entire segment to our positive contrastive set. For negative instances, we rely on the segment-scoring mechanism introduced in the previous subsection. Specifically, we consider the segments with high extraction probabilities ($p_i$) conditioning that they are non-gold as negative instances. We continue selecting negative instances until the number of selected negative segments matches the number of positive segments already selected. This selection process is dynamic and happens during training, meaning that the model is continually being challenged to distinguish between important and unimportant content. By incorporating both positive and negative instances, we ensure that our model is trained to focus on meaningful information, leading to better performance in downstream tasks. After identifying the positive and negative instances (i.e., segments), we feed them through a shared-weight decoder to generate a transformed representation for loss computations. 

\subsubsection{Computing contrastive loss}
To achieve our objective of maximizing the similarity between positive instances and their corresponding query embeddings while minimizing the similarity between negative instances and query embeddings, we opt to use the InfoNCE loss for contrastive learning, which is a popular contrastive learning technique introduced by \citet{Oord2018RepresentationLW} and has been proven to exhibit robust performance on QA tasks in recent studies~\cite{liu2021bridge,caciularu2022long,hu-etal-2022-momentum}.

A few studies have acknowledged the significance of feature transformation employing a small neural network for mapping representations to a space where contrastive loss is computed~\cite{Chen2020SimCTL,He_2020_CVPR}. Drawing inspiration from these works, and to address the challenge of decoder outputs being optimized for the token prediction task rather than semantic similarity within our framework, we utilize a multi-layer perceptron (MLP) coupled with Batch Normalization (BN)\cite{Ioffe2015BatchNA} and Rectified Linear Unit (ReLU) activation function\cite{Agarap2018DeepLU} to transform the decoder outputs (i.e., logits) into a new embedding space, as described below:

\begin{equation}
\begin{aligned}
h_i &= \texttt{ReLU}( \texttt{BN}( W_i d_i + b_i)) \\
s_i &= W_{j} h_i  + b_{j}
\end{aligned}
\end{equation}
in which $d_i$ is the decoder outputs (i.g., logits),  $W_i$, $W_j$, $b_i$, and $b_j$ are trainable parameters.

 We continue to use cosine similarity as the pairwise similarity function to measure the similarity between contrastive and query embeddings as follows:

\begin{equation}
\text{sim}(s_i, q) = \frac{\textbf{q}^\top \textbf{s}_i}{|\textbf{q}| |\textbf{s}_i|}
\end{equation}
where $\textbf{q}$ is the query embedding (i.e., summary generated from the entire input segments in our framework), $\textbf{s}_i$ is the contrastive instance (i.e., positive/negative) embedding, and $sim$ is the cosine similarity between the query and the contrastive instances. It is worth noting that since the output of the decoder network for $\textbf{q}$ and $\textbf{h}_i$ are token sequences, we compute the token-wise cosine similarity. We then define the InfoNCE contrastive loss as follows:

\begin{equation}
\mathcal{L}_{cont} = -\log \left(\frac{e^ {\text{sim}(s^+,q_i)/\tau}}{\displaystyle\sum_{s \in S} e^{\text{sim}(s,q)/\tau}}\right)
\end{equation}
where $sim(\cdot)$ denotes the similarity score between the query and the positive ($s^{+}$) and a contrastive ($s$) instances, $S$ is the set of contrastive instances, and $\tau$ is a temperature hyperparameter that controls the concentration of the probability distribution over the instances.

The InfoNCE loss encourages the model to learn embeddings that are more similar for positive instances and less similar for negative instances, effectively improving the model's ability to discriminate between important and unimportant content. The temperature parameter $\tau$ controls the sharpness of the distribution, with lower values of $\tau$ leading to a more focused distribution around the highest-scoring positive instance and higher values resulting in a smoother distribution across all instances. This parameter can be fine-tuned to strike a balance between focusing on the most relevant content and being robust to a diverse set of instances.

\subsection{Joint Training Objective}
We train our model by optimizing a joint objective that combines the generation loss, classification loss, and contrastive loss, introduced earlier as follows:

\begin{equation}
\mathcal{L} = \lambda_0\mathcal{L}_{\texttt{gen}} + \lambda_1 \mathcal{L}_{\texttt{cls}} + \lambda_2 \mathcal{L}_{\texttt{cont}}
\end{equation}
where $\lambda$ parameters balance the learning between the three tasks, and their sum is equal to 1.

\section{Experimental Setup}
\label{sec:setup}
In this section, we elaborate on the datasets, and baseline models that we have utilized in our experiments. We also provide training and implementation details. 

\subsection{Datasets}
\textbf{QMSum~\cite{zhong2021qmsum} } is a query-focused meeting summarization dataset consisting of 1,808 query-focused summaries extracted from 232 multi-turn meetings~\footnote{A ``multi-turn  dataset'' is a collection of conversational data that involves multiple rounds of dialogue between two or more participants such as those in the meetings. } across various domains, including product design, academia, and political committees. Additionally, the dataset includes annotations such as topic segmentations and highlighted text spans associated with reference summaries. The dataset is split into 1,257, 272, and 279 instances for training, validation, and testing, respectively. The length statistics represent the average source length of 9K tokens and summary length of 70 tokens. Given the widespread use of QMSum in prior research, we believe it serves as a valuable benchmark for comparison in our work.

\noindent \textbf{SQuALITY~\cite{wang-etal-2022-squality} }  is a collection of question-focused abstractive summarization data comprising 100 stories, 500 questions, and 2000 summaries. Each question in the dataset is accompanied by 4 reference summaries, written by trained writers who reviewed each other's work to ensure the data is of high quality. The dataset provides 39/25/36 (train/validation/test) splits, which is equivalent to 195/125/180 document-question pairs. The documents and summaries have a length of 5.2K and 237 tokens on average, respectively. 

\subsection{Baselines}
We experiment with different strong and state-of-the-art summarization models, which are outlined as follows. 

\begin{itemize}[leftmargin=10pt,label={-}]
    \item \textbf{Bart} is a Transformer-based encoder-decoder~\cite{Vaswani2017Att} model pretrained on a token infilling and sentence permutation pretraining objectives~\cite{Lewis2020BARTDS}. \bart netowrk has a maximum input length of 1024 tokens, hence the documents are truncated to fit this baseline.
    
    \item \textbf{Pegasus} is an abstractive summarizer that is pretrained using a task-specific objective for summarization~\cite{zhang2019pegasus}. Specifically, it is pretrained to predict the masked-out sentences as Gap Sentence Prediction (GSP) objective. The inputs are truncated to 2048 tokens to fit this baseline~\cite{wang-etal-2022-squality}.  

    \item \textbf{Bart+DPR} is an extract-then-summarize method amongst the baselines on SQuALITY dataset as proposed by \citet{wang-etal-2022-squality}. Unlike \bart, it first retrieves the sentences that are most relevant to the question and concatenates them to form the input to the abstractive summarizer. 

    \item \textbf{LED} is Longformer Encoder-Decoder model, amongst the abstractive summarization systems suited for long document summarization tasks~\cite{Beltagy2020LongformerTL}. This model modifies the conventional self-attention mechanism of Transformers architecture for a more efficient and robust scale-up to long documents.

    \item \textbf{Bart-LS} is an extension of Bart model for long document summarization, adapting it for long-sequence inputs. In particular, it replaces the full attention mechanism in Transformers with pooling-augmented block-wise attention, and pretrains the model with a masked-span prediction task with spans of varying lengths. 
    
    \item \textbf{DialogLM} is a pre-trained neural model for understanding and summarizing long dialogues. It uses an encoder-decoder architecture and a window-based denoising pre-training task to equip the model with the ability to reconstruct noisy dialogue windows.

    \item \textbf{SegEnc} is a state-of-the-art abstractive summarizer, tailored to address query-focused long-input documents. It divides the input into fixed-length segments, encodes them separately, and then enables the decoder to jointly attend to these segments. Two different configurations of the \segenc model are employed: (1) the default checkpoint, which is the \texttt{bart-large} architecture; and (2) the \textit{Wikisum Pre-Finetuned} model (aka \segenc-W), which performs pre-finetuning of the \segenc model on the Wikisum dataset. 
    
    \item \textbf{Socratic} is a pre-training framework that uses a question-driven objective specifically designed for controllability in summarization tasks~\cite{Pagnoni2022SocraticPQ}. This framework trains a model to generate and respond to relevant questions (i.e., \textit{ask \& answer}) in a given document and achieves the SOTA on QMSum and SQuALITY datasets. Socratic is finetuned on \segenc summarization system. We also include a version of \pegasus that is pretrained on \textit{Book3} pre-training dataset, and fine-tuned using \segenc approach. 

    \item \textbf{TUQFS} is a query-aware approach that employs joint modeling of tokens and utterances through Token-Utterance Attention~\cite{Liu2023TUQFS}. This technique integrates both token-level and utterance-level query relevance into the generation process via an attention mechanism. In our experiments, the TUQFS model is considered one of the state-of-the-art methods for the QMSum dataset.
    
\end{itemize}

\subsection{Training and implementation details}
Our method is implemented using Huggingface Transformers. For the model's hyperparameters, we set a learning rate of 5e-5 with a weight decay of 0.01 and train the model for 10 epochs, with validation conducted at the end of each epoch. We select the checkpoint that achieves the highest mean \textsc{Rouge} scores for inference time, a strategy similar to \segenc~\cite{Vig2022SEGENC}. To compute the contrastive loss, we tune the $\tau$ parameter from the set of \{0.2, 0.4, 0.6, 0.8\} and fix it at 0.6 and 0.8 for the QMSum and SQuALITY datasets, respectively. We tried different values of $\lambda$ in joint learning and fixed them on ($\lambda_0$=0.6, $\lambda_1$=0.2, $\lambda_2$=0.2) for both datasets.

In the SQuALITY dataset, unlike the QMSum dataset, there are no human-annotated ground-truth spans for each query-summary pair. To provide supervised span labels, we adopt a method similar to that used in ~\cite{song-etal-2022-grounded} and label the input segments based on the word bigram overlap between the segment and summary. If a segment has six or more common bigrams with the summary, it is labeled as positive; otherwise, it is labeled as negative.

\section{Experiments}
In this section, we present the automatic evaluation results, as well as the findings from a human study conducted to compare the system-generated outputs against each other.

\subsection{Automatic results}
The performance of various summarization models on the QMSum and SQuALITY benchmarks was evaluated using the \textsc{Rouge} and \textsc{BertScore} metrics, as displayed in Tables \ref{tab:qmsum-auto} and \ref{tab:squality-auto}, respectively. Our proposed method, \ours, outperforms the majority of the baselines across all metrics in the QMSum benchmark. When compared to the \socratic Pret.~\cite{Pagnoni2022SocraticPQ} as the strongest baseline, our system achieves improvements of +0.36 (1\%) \ro, +0.52 (1.5\%) \rl, and +0.09 (0.1\%) \textsc{BertScore} points (relative improvements). Although our method falls short by 0.24 points in \rt compared to the \socratic Pret., it is crucial to emphasize that our approach outperforms the state-of-the-art  \socratic Pret. model.

In terms of computational cost, our proposed \ours model offers a distinct advantage over the \socratic model. Specifically, our model leverages a contrastive learning strategy that obviates the need for extensive pre-training, which typically requires a significantly larger dataset and consequently, leads to increased computational overhead. The \socratic Pret. 1M model and \socratic Pret., for instance, underwent an extensive pre-training phase involving 1 million, and 30 million instances from the \textit{Book3} collection. In contrast, our \ours model only 
employs contrastive learning during the fine-tuning stage, which uses a significantly smaller dataset. This difference in approach has a direct effect on computational efficiency: the drastically reduced dataset size in our model's training phase reduces the computational cost 
significantly. Despite this reduction in training complexity, our model was able to achieve improved or at least comparable results across all metrics when compared to \socratic. These results not only illustrate the efficiency of our model but also open a new perspective on how to construct more computationally efficient models for Query-focused Summarization tasks.

\begin{table}[t]
\centering
\begin{tabular}{lcccc}
\toprule
 & \textbf{RG-1} & \textbf{RG-2} & \textbf{RG-L} & \textbf{BS} \\
\midrule
Bart~\cite{Lewis2020BARTDS} & 29.20 & 6.37 & 25.49 & - \\
Bart-LS~\cite{xiong2022adapting} & 37.90 & 12.10 & 33.10 & - \\
DialogLM~\cite{Zhong2021DialogLMPM} & 34.00 & 9.20 & 30.00 & - \\
TUQFS~\cite{Liu2023TUQFS} & 37.99 & 13.66 & 33.36 & - \\
\midrule 
SegEnc~\cite{Vig2022SEGENC} & 37.05 & 13.03 & 32.62 & 87.44 \\
\hspace{.1em} \textit{\small+ Wikisum Pre-Finetuned}~\cite{Vig2022SEGENC} \normalsize& 37.80 & 13.43 & 33.38 & - \\
\hspace{.01em} \small+ \socratic Pret. 1M~\cite{Pagnoni2022SocraticPQ} \normalsize & 37.46 & 13.32 & 32.79 & 87.54 \\
\hspace{.01em} \small + \socratic Pret.~\cite{Pagnoni2022SocraticPQ} \normalsize & 38.06 & \textbf{13.74} & 33.51 & 87.63 \\
\midrule
\ours (ours) &\textbf{ 38.42} & {13.50} & \textbf{34.03} & \textbf{87.72} \\
\bottomrule
\end{tabular}
\caption{Performance of various summarization systems in terms of \textsc{Rouge} (RG) and \textsc{BertScore} (i.e., BS) on QMSum dataset. The top and middle boxes are reported performances from previous works~\cite{zhong2021qmsum,Vig2022SEGENC,Pagnoni2022SocraticPQ}. While \socratic Pret. is the strongest baseline, it was not released at the time of writing this paper; hence, we reproduced \segenc + \textit{Wikisum Pre-Finetuned} for human evaluation purposes.}
\label{tab:qmsum-auto}
\end{table}

\begin{table}[t]
\centering
\begin{tabular}{@{}lcccc@{}}
\toprule
 & \textbf{RG-1} & \textbf{RG-2} & \textbf{RG-L} & \textbf{BS}\\
\midrule
LED~\cite{Beltagy2020LongformerTL} & 27.7 & 5.9 & 17.7 & - \\
Pegasus~\cite{zhang2019pegasus} & 38.2 & 9.0 & 20.2 & - \\
Bart~\cite{Lewis2020BARTDS} & 40.2 & 10.4 & 20.8 & - \\
Bart + DPR~\cite{wang-etal-2022-squality} & 41.5 & 11.4 & 21.0 & - \\
\midrule
SegEnc~\cite{Vig2022SEGENC} & 45.68 & 14.51 & 22.47 & 85.86 \\
\hspace{.2em}+ \pegasus Pret.~\cite{Pagnoni2022SocraticPQ} & 45.78 & 14.43 & 22.90 & 85.94 \\
\hspace{.2em}+ \socratic Pret.~\cite{Pagnoni2022SocraticPQ} & 46.31 & 14.80 & 22.76 & 86.04 \\
\midrule

\ours (ours) & 45.76 & 14.27 & \textbf{24.14} & \textbf{86.07} \\
\bottomrule
\end{tabular}
\caption{Performance of various summarization systems in terms of \textsc{Rouge} (i.e., RG) and \textsc{BertScore} (i.e., BS) metrics on SQuALITY benchmark. The reported results are directly taken from the corresponding papers~\cite{wang-etal-2022-squality,Pagnoni2022SocraticPQ}.}
\label{tab:squality-auto}
\end{table}

Similarly, the \rl score of our proposed \ours model demonstrates a significant improvement compared to the SOTA baselines on the SQuALITY dataset (Table \ref{tab:squality-auto}), resulting in an increase of 1.24 (5\%) points (relative improvements). However, there is a relatively slighter decrease of 0.55 points (\ro) and 0.53 (\rt) when compared to the best baseline results. We hypothesize that the relatively lower scores in \ro and \rt can be attributed to the absence of gold spans, which were provided by humans in the QMSum experiments, used in the training of our \ours model. This finding suggests that there is room for improvement by employing more sophisticated techniques to enhance positive segment labeling, potentially leading to better performance in the \ro and \rt metrics.


\begin{figure}
    \centering
    \includegraphics[scale=0.36]{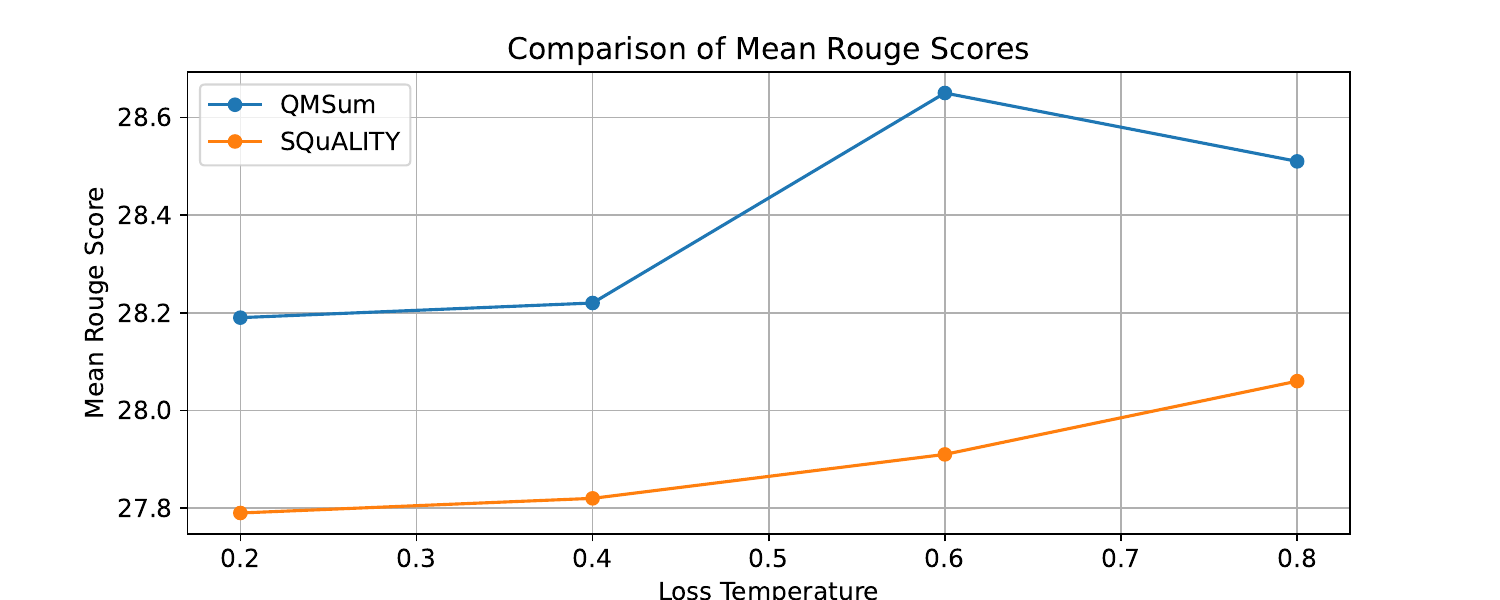}
    \caption{The effect of temperature parameter (infoNCE loss) to the overall performance of the model over two datasets.}
    \label{fig:enter-label}
\end{figure}

The impact of the temperature parameter (infoNCE loss) on system performance is demonstrated in Figure. \ref{fig:enter-label}. As depicted, the mean \rouge scores for both QMSum and SQuALITY datasets rise with an increase in the temperature parameter. In the case of the QMSum dataset, the score noticeably ascends from 0.4 to 0.6, then experiences a minor decline at 0.8, thereby suggesting that the most effective parameter value could be approximately 0.6. In contrast, the SQuALITY dataset displays a steady, though marginal, augmentation in the score with increasing parameter values, hinting at a more linear correlation. Generally, across all temperature parameters, the QMSum dataset exhibits a slightly superior mean ROUGE score compared to the SQuALITY dataset.

\subsection{Human study}
Many previous studies have acknowledged the limitations of automatic evaluation metrics and their correlation with human judgments. To gain a better understanding of system-generated summary qualities and provide a basis for error analysis, we aim to conduct a human study on randomly sampled summaries. This approach will allow us to obtain insights into the strengths and weaknesses of our proposed \ours model and identify areas where it can be improved. By comparing the results of the human study with the automatic evaluation metrics, we can gain a deeper understanding of the performance of our model and make informed decisions on how to optimize it further. The results of this human study will complement the experimental findings presented in this paper and contribute to a more comprehensive evaluation of our proposed summarization model.

In order to conduct the human study, we randomly selected a list of 50 meeting-query instances from the QMSum dataset and provided the human annotator with summaries generated by three different methods: human-written, \segenc-W generated, and \ours generated. Due to the typically lengthy nature of meeting transcripts, which  often exceeded 9K tokens in our study, we only provided the gold spans relevant to the given query during our evaluation process. This allowed the annotator to focus specifically on the information that was most important and directly related to the query, without being overwhelmed by extraneous information. To avoid any bias, we shuffled the order in which the summaries were presented for evaluation. We then defined three qualitative metrics for the evaluation process, which are listed below, and  each case was scored on a Likert scale ranging from 1 (best) to 5 (worst).

\begin{itemize}[leftmargin=20pt,label={-}]
    \item\textbf{Fluency: } This metric measures the extent to which the summary is fully understandable (score of 5) or completely gibberish (score of 1); 
    
    \item\textbf{Relevance: } This metric quantifies how well the summary (even short responses within it) aligns with the given query, with a score of 5 for high relevance and 1 for low~\footnote{Note that for the relevance metric, we do not consider the information within the ground-truth summary, but rather  focusing solely on assessing the quality of the summary based on its relevance to the query at hand.};
    
    \item\textbf{Faithfulness:} This metric measures the extent to which the materials produced in the summary are supported (score of 5) or not supported (score of 1) within the meeting.
\end{itemize}

\begin{table}[t]
\centering
\label{tab:human-eval}
\begin{tabular}{@{}lccc@{}}
\toprule
\multicolumn{1}{c}{\textbf{Method}} & \textbf{Fluency} & \textbf{Relevance} & \textbf{Faithfulness} \\ \midrule
Human-written & 4.34 & 4.72 & 4.85 \\
\midrule 
\segenc-W~\cite{Vig2022SEGENC} & 4.22 & 3.84 & \textbf{3.52} \\
\ours (ours) & \textbf{4.23} & \textbf{3.96} & 3.42 \\ \bottomrule
\end{tabular}
\caption{Results of the human study on 50 evaluation samples from the QMSum dataset, comparing system-generated summaries with human-written ones. While there is parity between human performance and system performances, our proposed \ours method improves the relevance metric over the \segenc-W baseline, without sacrificing fluency.}
\label{tab:human-qmsum}
\end{table}

The results of the human study presented in Table \ref{tab:human-qmsum} show that our proposed \ours method achieves a comparable level of fluency to human-written summaries, which is justifiable considering the current state-of-the-art abstractive summarization systems given their strong pre-training ability. However, there is a clear gap between the performance of systems and that of human-written summaries on the qualitative metrics. Despite this, our \ours method outperforms the \segenc-W baseline, particularly on relevance, where it achieves an average score of 3.96 compared to 3.84 for \segenc-W. This demonstrates that our model is better at capturing the most important and relevant information in the meeting transcripts and summarizing it effectively, without sacrificing fluency over the baseline system. We also see a gap between the relevance metric of human-written summaries and system-generated summaries, which might be due to the inclusion of irrelevant spans of information in the system-generated summaries compared to the human-written summaries that have been written from the gold spans. Despite this, our \ours method shows promise in improving the relevance metric. However, there is still a need for further research to improve the relevance and faithfulness metric, especially in domain-specific contexts like meeting transcripts. The observed gap in faithfulness between human performance and summarization systems, as also pointed out in previous studies~\cite{zhong2021qmsum,Vig2022SEGENC}, might be attributed to the challenges of understanding and representing natural language nuances in domain-specific contexts like meeting transcripts.

\begin{figure*}[t]
    \centering
    \includegraphics[scale=0.34]{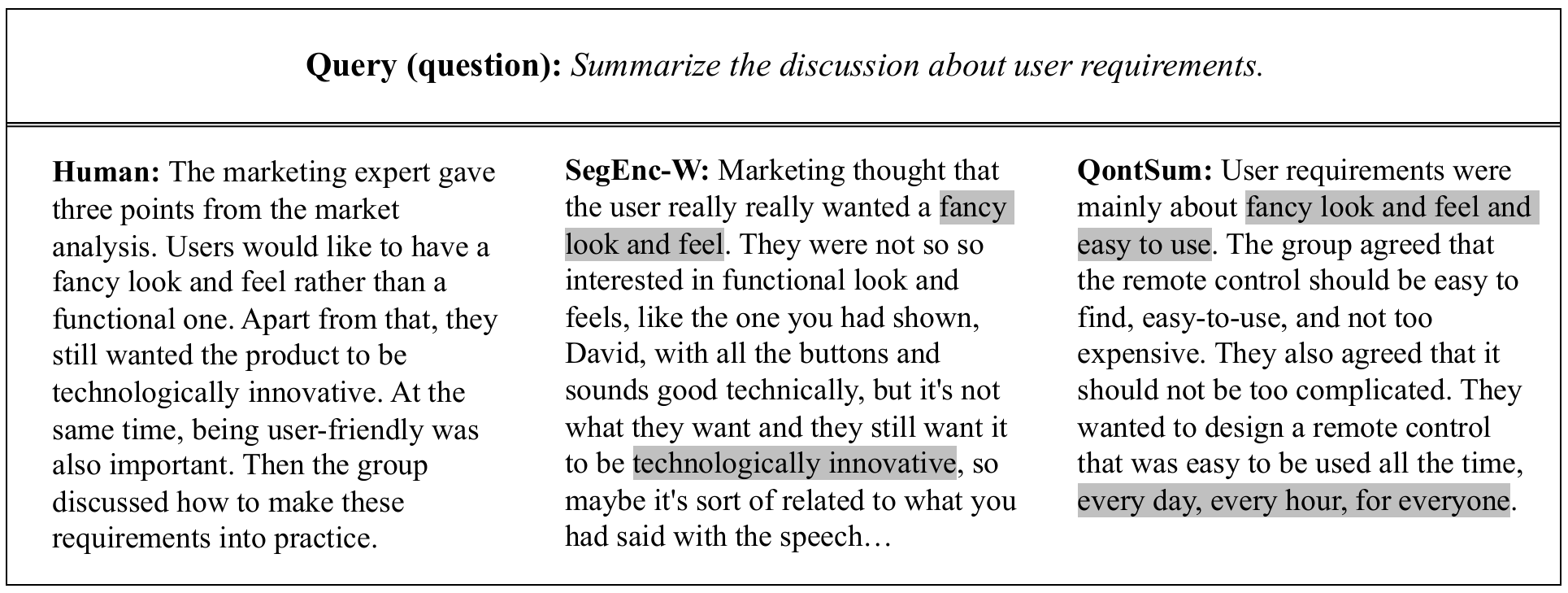}
    \caption{A sample case from the error analysis study. As demonstrated, while \ours was able to capture a few relevant pieces of information (highlighted in the summary), it failed to mention critical points, such as the product's ``technological innovation''. Additionally, the sample case exposes a common error in our generated summaries, which is the repetition of words or phrases. In this case, the product's ``easy-to-use'' nature is referenced multiple times in the summary, leading to a lack of fluency. On the other hand, while the baseline method was able to capture useful information, it suffered from a \textit{transcripts copying} error in the final truncated sentence. These findings indicate that there is still room for improvement in both our proposed method and the baseline method. Further research could explore ways to address these limitations and improve the quality of the generated summaries. }
    \label{fig:an2}
\end{figure*}

\begin{figure*}[t]
    \centering
    \includegraphics[scale=0.34]{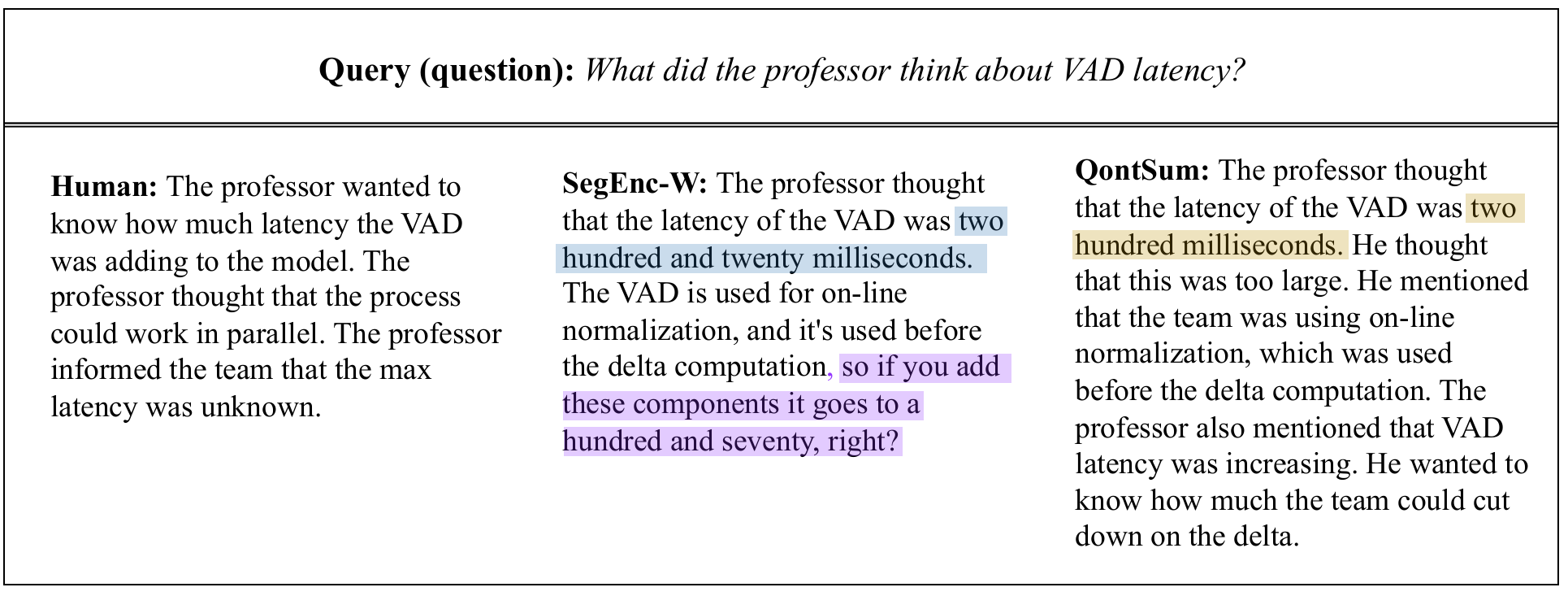}
    \caption{A sample case from the error analysis study. As shown, the \ours-generated summary has taken a partial phrase (i.e., two hundred) from the transcript (golden highlight), while \segenc-W could bring the complete phrase (blue highlight), although the material does not appear in the Human-written summary. The purple highlight represents instances of \textit{transcript copying}, which were observed during the evaluation process and affected the fluency of the generated summaries. This case highlights one of the limitations of our proposed approach and suggests that future research could explore ways to address this issue to further improve the quality of the generated summaries}
    \label{fig:an1}
\end{figure*}

\subsection{Error analysis}

In the following, we present our findings on the types of errors observed affecting each qualitative metric, their causes, and potential solutions to address the detected issues.

\subsubsection{\textbf{Fluency.} }
When evaluating our system on the {fluency} metric, we found that its performance was generally comparable to that of the human-written and baseline systems. However, in cases where our system underperformed the other systems, we observed a couple of common errors, particularly \textit{information repetition} (Figure. \ref{fig:an2}), \textit{incoherence}, \textit{trasncripts copying} (Figure. \ref{fig:an1}), and \textit{speaker mix-up}, particularly on QMSum benchmark. 

We note that segmenting the input documents could have a negative impact on fluency in summarization. Specifically, since the segmentation is performed on fixed-length chunks, it may cause boundary sentences to be split across segments, potentially affecting the coherence of the system summary. We further observed that the current state-of-the-art abstractive summarization models, including our proposed system, face significant challenges in summarizing multi-turn meeting transcripts. 
    
These challenges stem from the complexity of the task, which involves capturing the nuances of human conversation, understanding the speaker's intent and context, and producing a coherent summary that reflects the key takeaways from the discussion. Addressing these challenges will require further research and the development of novel techniques that can effectively handle the complexity of multi-turn summarization tasks such as the works done by ~\cite{zhong2021qmsum,Zhong2021DialogLMPM}.

\subsubsection{\textbf{Relevance}}
When evaluating the system's performance in terms of {relevance}, we found that our proposed system outperformed the baseline but still fell short of human parity. In investigating the common reasons contributing to the errors observed in the underperformed cases, we identified several factors. 
    
First, we found that the system sometimes struggled to accurately identify the most important information relevant to the query, leading to summaries that were less informative or missed critical details. This issue was more prevalent on the SQuALITY dataset, where no span annotations were provided, making it more challenging for the system to identify the relevant information. However, on the QMSum benchmark, where high-quality human-annotated text spans were provided, we observed fewer instances of this problem. This suggests that the integration of enhanced explicit supervision or utilization of more advanced labeling techniques may potentially contribute to the system's performance for selecting pertinent information, thereby generating more relevant summaries. 
    
Furthermore, We also observed that the system's ability to capture the nuances of the query was sometimes limited, especially in cases where the query was too broad or too specific. For instance, open-ended queries such as \textit{``Summarize the whole meeting''} lack specificity and can be interpreted in multiple ways, making it challenging for the system to determine which information is most relevant to include in the summary, particularly in longer documents. On the other hand, extremely specific queries, such as \textit{``What did Marketing think of the incorporation of current fashion trends in the prototype when making a simulation market evaluation of the new remote control?'}' can also pose challenges for the system. In such cases, the system may struggle to accurately interpret the underlying meaning of the query, leading to summaries that may not fully address the information needs.

Finally, we also noted that the choice of input document could significantly impact the system's performance, with some documents being more challenging for the system to summarize effectively. For instance, documents that contain multi-turn dialogues or include idiomatic expressions that are often made during meetings can pose significant challenges for the system. Similarly, the use of figurative language or domain-specific terminology can also impact the system's ability to accurately interpret the underlying meaning of the input document. To address these issues, we suggest further exploration of techniques for improving the system's ability to identify and select relevant information, such as using more advanced supervision signals (e.g., improving the semi-supervised labeling mechanism) or incorporating external knowledge sources to learn in-domain terminologies more effectively.

\subsubsection{\textbf{Faithfulness. }} 

Faithfulness is a crucial aspect of summarization, as it ensures that the summary accurately reflects the key information from the input document. However, it has been identified as a significant challenge for long-input summarization tasks, including Query-Focused summarization. Prior works have also reported similar challenges in achieving faithful summaries~\cite{zhong2021qmsum,Zhong2021DialogLMPM,Vig2022SEGENC} that accurately capture the relevant information from the input document and address the information needs.

In our error analysis, we observed several instances where the proposed model produced summaries that were not faithful to the input document. Specifically, we observed a common error where the generated summary was written from multiple segments that all did not align with the entire query. This could be attributed to the model's inability to effectively filter out irrelevant information or to correctly interpret the underlying meaning of the query. For example, the model may have included information that was only tangentially related to the query, but not directly relevant to the user's information needs. For instance, a query such as \textit{``Did the group think the remote control was easy to use when discussing evaluation criteria of the remote control?''} might be asked. However, when it comes to interpreting the query meaning, the model might  put more focus on only \textit{``being easy to use''} and include segments that discuss all items that should be easy to use without considering the specific context and information within the query, such as the group's discussion on the remote control's evaluation criteria. In such cases, the model mixes up different items from the attended segments and generates a summary that is not faithful to the input. 

Another error we observed was the omission of critical details or information that was relevant to the query, resulting in incomplete or inaccurate summaries, as shown in Figure. \ref{fig:an1}. This could be due to the model's  limitations in accurately interpreting and analyzing the contextual information in the input document. 
Another error was that the model exhibited a deficiency in tracking specific items that were frequently referenced during the meeting. This resulted in a failure to accurately capture the final decision or outcome that was made regarding a specific discussion. In other words, the model demonstrated an inability to consistently comprehend and retain crucial information regarding the frequently mentioned discussions on specific items. This type of error can have significant consequences, especially in contexts where the mentioned items play a critical role in decision-making.

To address these errors and improve the faithfulness of the system, we suggest exploring techniques such as integrating fact-checking mechanisms. Additionally, further research could be conducted to develop more advanced natural language understanding techniques such as reinforcement learning with faithfulness-focused rewards, which encourage the model to increase the faithfulness of the generated summaries. 
\section{Conclusion}

This study proposes a novel approach for query-focused summarization (QFS) that employs contrastive learning to enhance the relevance of the summary to the given query. The proposed method utilizes the relevant segments of the document as positive instances and high-scored non-gold segments during the training process as negative instances for generating contrastive samples. Specifically, after identifying contrastive segments, they are fed into the abstractive summarizer for generating summaries for each contrastive instance, which then contribute to the computation of the contrastive loss (i.e., InfoNCE loss function). The entire network is optimized using a joint loss that combines generation, classification, and contrastive losses with balancing hyperparameters. Experimental results indicate that the proposed method outperforms existing state-of-the-art techniques and achieves a new SOTA (with 1\% and 1.5\% improvements over \ro and \rt, respectively on QMSum; and 1.24 (5\%) \rl on SQuALITY benchmarks as compared to the previous SOTA) or comparable performance with reduced computational overhead without further large-scale pretraining. Furthermore, a human study analysis demonstrates the effectiveness of the approach in terms of relevance, without sacrificing fluency. The conducted error analysis further provides insights into the current limitations and future research directions for QFS. 
The study's contribution adds to the growing body of research on natural language generation and has the potential to advance the state-of-the-art in QFS. Overall, the proposed method represents a  step forward in addressing the challenge of QFS in Gen-IR, and it is hoped that it will inspire further research in this area.


\bibliographystyle{ACM-Reference-Format}
\bibliography{samples/sample-base}

\end{document}